\documentclass{llncs}
\usepackage[utf8]{inputenc}
\usepackage{numprint}
\usepackage{todonotes}
\usepackage{hyperref}
\usepackage{microtype}
\usepackage{listings}
\usepackage{tabto}
\usepackage{wrapfig}
\usepackage{aliascnt}
\usepackage{times}
\usepackage{helvet}
\usepackage{courier}
\usepackage{verbatim}
\usepackage{booktabs}
\usepackage{siunitx}
\usepackage{graphicx}
\usepackage[english]{babel}
\usepackage{comment}
\usepackage{amsmath}
\usepackage{balance}
\usepackage{amssymb}
\usepackage{epstopdf}
\usepackage{tabularx}
\usepackage{footmisc}
\usepackage[T1]{fontenc}
\usepackage[scaled=0.85]{beramono}
\usepackage{algorithm}
\usepackage{algorithmic}
\usepackage{lipsum}
\usepackage{paralist}
\usepackage{multirow}
\newcommand{\furl}[1]{\footnote{\scriptsize \url{#1}}}

\lstdefinelanguage{scala}{
  morekeywords={abstract,case,catch,class,def,%
    do,else,extends,false,final,finally,%
    for,if,implicit,import,match,mixin,%
    new,null,object,override,package,%
    private,protected,requires,return,sealed,%
    super,this,throw,trait,true,try,%
    type,val,var,while,with,yield},
  otherkeywords={=>,<-,<\%,<:,>:,\#,@},
  sensitive=true,
  morecomment=[l]{//},
  morecomment=[n]{/*}{*/},
  morestring=[b]",
  morestring=[b]',
  morestring=[b]"""
}

\newcommand{\printfnsymbol}[1]{%
  \textsuperscript{{*}}%
}

\title{VANiLLa : Verbalized Answers in \newline Natural Language at Large Scale}

\author{Debanjali Biswas\inst{1} \thanks{Equal contribution}
\and  Mohnish Dubey\inst{2,3} \printfnsymbol{1}
\and \newline
Md Rashad Al Hasan Rony\inst{3}
 \and  Jens Lehmann\inst{2,3}}
\authorrunning{ Biswas, Dubey, Rony, Lehmann}
\institute{
Department of Language Science and Technology, Saarland University, Germany \newline
\email{dbiswas@coli.uni-saarland.de}
\and
Smart Data Analytics Group (SDA), University of Bonn, Germany \newline
\email{\{dubey, jens.lehmann\}@cs.uni-bonn.de}
\and
Fraunhofer IAIS, Dresden, Germany
\newline
\email{\{mohnish.dubey,rashad.rony,jens.lehmann\}@iais.fraunhofer.de}
}

\begin{document}
\maketitle
\vspace{-10pt}
\begin{abstract}
{In the last years, there have been significant developments in the area of Question Answering over Knowledge Graphs (KGQA). Despite all the notable advancements, current KGQA datasets only provide the answers as the direct output result of the formal query, rather than full sentences incorporating question context. For achieving coherent answers sentence with the question's vocabulary,  template-based verbalization so are usually employed for a better representation of answers, which in turn require extensive expert intervention. Thus, making way for machine learning approaches; however, there is a scarcity of datasets that empower machine learning models in this area. Hence, we provide the VANiLLa dataset which aims at reducing this gap by offering answers in natural language sentences. The answer sentences in this dataset are syntactically and semantically closer to the question than to the triple fact. Our dataset consists of over 100k simple questions adapted from the CSQA and SimpleQuestionsWikidata datasets and generated using a semi-automatic framework. We also present results of training our dataset on multiple baseline models adapted from current state-of-the-art Natural Language Generation (NLG) architectures. We believe that this dataset will allow researchers to focus on finding suitable methodologies and architectures for answer verbalization.


}
\end{abstract}
\vspace{-10pt}
\noindent \textbf{Resource Type}: Dataset \\
\textbf{Website and documentation}: \url{https://sda.tech/projects/vanilla/}\\
\textbf{Permanent URL:} \url{https://figshare.com/articles/Vanilla_dataset/12360743}

\section{Introduction}

The goal of Question Answering over Knowledge Graphs (KGQA) is to retrieve the answer from a knowledge graph (KG) for a given natural language question. 
A conventional solution to KGQA is to translate natural language questions to a formal query, which then returns the intended answer when executed over the KG. 
The primary KGQA data sets, such as SimpleQuestion~\cite{bordes2015largescale}, WebQuestions~\cite{yih-etal-2016-value}, LC-QuAD~\cite{DBLP:conf/semweb/TrivediMDL17} and KGQA systems~\cite{HoffnerWMULN17} consider the formal query response as the final answer to the question.
KGQA challenges such as QALD~\cite{inproceedings3} consider the formal query or their response for evaluations.
However, these responses might leave the final end-user unsatisfied with the answer representation. 

\vspace{-10pt}
\begin{figure}[h]
\centering
\includegraphics[width=\textwidth]{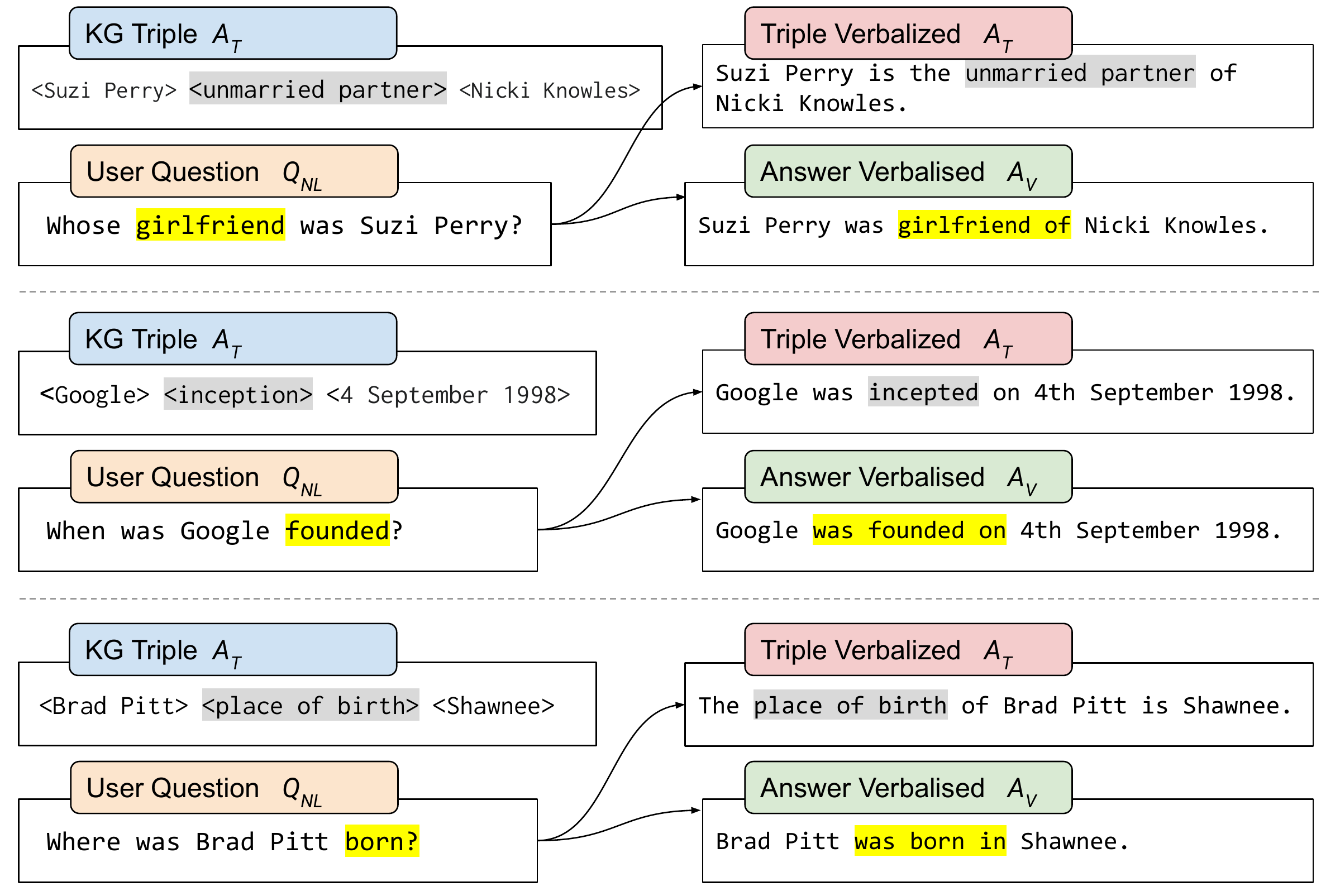}
\vspace{-10pt}
\caption{Examples from Vanilla Dataset}
\label{fig1}
\end{figure}
\vspace{-10pt}

For example,for the question $Q_{NL}$ ``\textit{Whose girlfriend was Suzi Perry?}'', a KGQA system could answer the question based on the KG triple $T$ \textit{<Suzi Perry> <unmarried partner> <Nicki Knowles>}.
The formal query (SPARQL) $Q_{NL}$ for this question could be \texttt{`SELECT ?ans WHERE \{<Suzi Perry> <unmarried partner> ?ans\}'}
\footnote{equivalent wikidata SPARQL is \textit{`SELECT ?ans WHERE \{ wd:Q7651082 wdt:P451 ?ans\}}'}
, and the final answer retrieved $A_R$ would be \textit{`Nicki Knowles'}. 
If Question Answering system further verbalizes the answers to \textit{`Suzi Perry was girlfriend of Nicki Knowles.
`} by incorporating $Q_{NL}$ context, the end-user will feel more satisfied with answer as a sentence when compared to receive the answer just as a date. 


A simple solution for verbalising answers is to design a template-based solution. However, the template solution can be expensive as it requires field experts to create the template. Moreover, the coverage is limited to queries matching the provided templates and the templates may be too inflexible to take particular grammar constructs, aggregate queries and context into account. In particular in multi-turn dialogues, templates can be repetitive and lead to unnatural conversations.

For this reason, machine learning techniques have been investigated for answer verbalization. These techniques, in turn, depend primarily on large-scale training data. At present, the datasets that can be applied for answer verbalization tasks are mostly RDF verbalization datasets like WebNLG~\cite{colin-etal-2016-webnlg}. 
Whereas in  VQuAnDa dataset~\cite{kacupaj2020vquanda} (we discuss this dataset in detail in section \ref{Sec:Litrature}), they verbalized the answer based on the SPARQL query.
Using such RDF/Query verbalization technique, we can verbalize the triple, but they might be not apt for the answer representation as they do not take the question context into account. 
As an example, consider the upper question in Figure~\ref{fig1}. In this case, the triple verbalization $A_T$ `Suzi Perry is the unmarried partner of Nicki Knowles.' is less suitable than the human like response "Suzi Perry was girlfriend of Nicki Knowles", which uses the word "girlfriend" from the user query.
In order to spur research into such context-driven answer verbalizations for KGQA, we present the VANiLLa: `Verbalized Answers in Natural Language at Large scale` where the verbalized answer sentences are generated by using the context of question to encourage human-like responses.

The following are key contributions of this work:
\vspace{-1pt}
\begin{itemize}
	\item Provision of a large-scale dataset of over 100k questions along with their corresponding verbalized answers.
	\item We present a cost-efficient semi-automatic framework for generating large-scale answer verbalization datasets, which was also used to generate VANiLLa.
	\item We provide multiple baseline models based on the conventional sequence-to-sequence framework as reference for future research. 
	\item We also provide empirical proof that answer verbalizations are better suited response and provide more context compared to RDF triple verbalizations.
\end{itemize}

This paper is organised into the following sections: \begin{inparaenum}[(1)] \setcounter{enumi}{1}
\item Problem Statement, 
\item Dataset Generation Workflow, 
\item Characteristics of the Dataset, 
\item Evaluation results,
\item Potential impact of this datset along with some Related Work, and
\item Conclusion and Future Work.
\end{inparaenum}


\section{Problem Statement} 

\begin{figure}[t]
\centering
\includegraphics[width=0.88\textwidth]{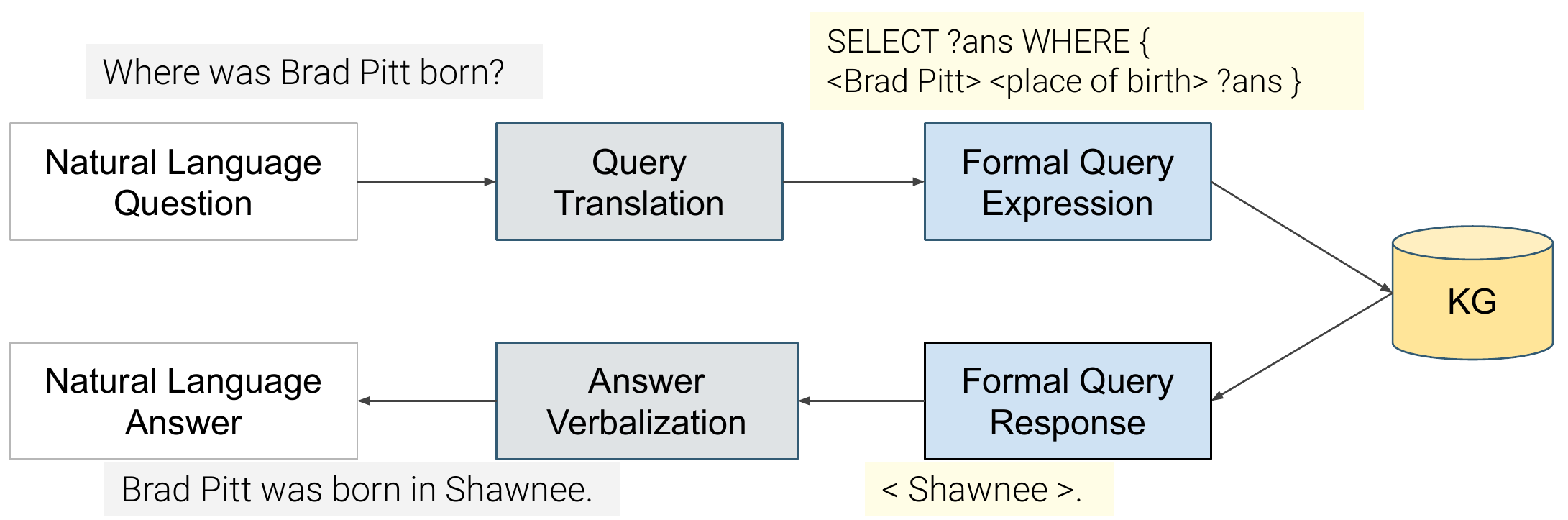}
\vspace{-5pt}
\caption{The QA pipeline for generating the answer in natural language} \label{Qa}
\vspace{-5pt}
\end{figure}

Formally, KGQA solutions are based on the semantic matching of a Natural Language Question $Q_{NL}$ to KG triples. The goal is usually to generate Formal expression $Q_{FL}$, such that  $Q_{NL}$ and $Q_{FL}$ have an equivalent semantic interpretation. 
Results retrieved through a formal query (often SPARQL in KGQA) are often either entities, literals or boolean values (which is a special case of a literal that we treat separately) or a list of the previous items.
Giving these as response directly to a user question is sufficient to answer the question. However, the responses are often not human friendly and it is difficult for a user to assess whether the question was correctly understood.
Thus, the need for verbalizing answers in natural language in two fold: 
\begin{inparaenum}[(i)]
    \item to provide some context of the question in the answer to enable users to verify the provided answer~\cite{Diefenbach2017SPARQLtoUserDT}, and
    \item to mimic natural human conversations.
\end{inparaenum}
From this perspective, an extended model to generate the final representation of answer in the QA pipeline is recommended as shown in Figure: \ref{Qa}. 

In this model, given a language question $Q_{NL}$ KGQA aims to retrieve a natural language answer $A_{NL}$ to model a transformation function $f_{kgqa}$ such that :
\[
 \centerline{ $f_{kgqa} : Q_{NL} \mapsto A_{NL}$ }
 \]
 \[
 \centerline{ $(Q_{NL})^{C_q} \equiv (A_{NL})^{C_a}$ } 
 \]
 \\
$C_q$ is the linguistic and semantic context of the natural language question\\
$C_a$ is the linguistic and semantic context of the natural language answer\\

Within the scope of this paper, we do not retrieve an answer from a KG for a given question. 
Rather, we focus on verbalizing the answers retrieved through the $Q_{FL}$ corresponding to the $Q_{NL}$. 
We describe this model as follows:

Given a user question $Q_{NL}$ and an answer retrieved $A_R$ by a QA system using $Q_{FL}$. The task of the model $f$ is to generate verbalized answer $A_V$ such that the context of the question is preserved and we can obtained a more human-like response. Formally,
\begin{equation}
    A_V = f(Q_{NL},A_R)
\end{equation}

We further introduce our hypothesis regarding the semantic closeness of natural language question(input to KGQA) and natural language answer(output of KGQA).
\subsubsection{Hypothesis H1:} \label{hyp}
For a given question $Q_{NL}$ and the retrieved answer $A_R$ from KG triples $T$, the verbalized answer $A_V$ with question context is semantically and syntactically closer to the question $Q_{NL}$ in comparison with that of the KG triple verbalization $A_T$ and $Q_{NL}$. Formally,
\begin{equation}\label{eq:2}
    Sim(A_V, Q_{NL}) > Sim(A_T, Q_{NL})
\end{equation}
Similarly, for a given question $Q_{NL}$ and the retrieved answer $A_R$ from KG triples $T$, the RDF verbalization $A_T$ is semantically and syntactically closer to the triple $T$ in comparison with that of the answer verbalization $A_V$ and $T$. Formally,
\begin{equation}
    Sim(A_T, T) > Sim(A_V, T)
\end{equation}
where, $Sim$ is the similarity measure. 

 For example, in context of equation \ref{eq:2}, Similarity ($Sim$) between `\textit{Brad Pitt was born in Shawnee}' and `\textit{Where was Brad Pitt born?}' is greater then Similarity between \textit{`The place of birth of Brad Pitt is Shawnee.'} and \textit{`Where was Brad Pitt born?'}.
We empirically prove this hypothesis with Experiment 1 in Section: \ref{Sec:Exp1}.


\section{Dataset Design and Characteristics}
\subsection{Dataset Generation Workflow}
\begin{figure} [h]
\includegraphics[width=\textwidth]{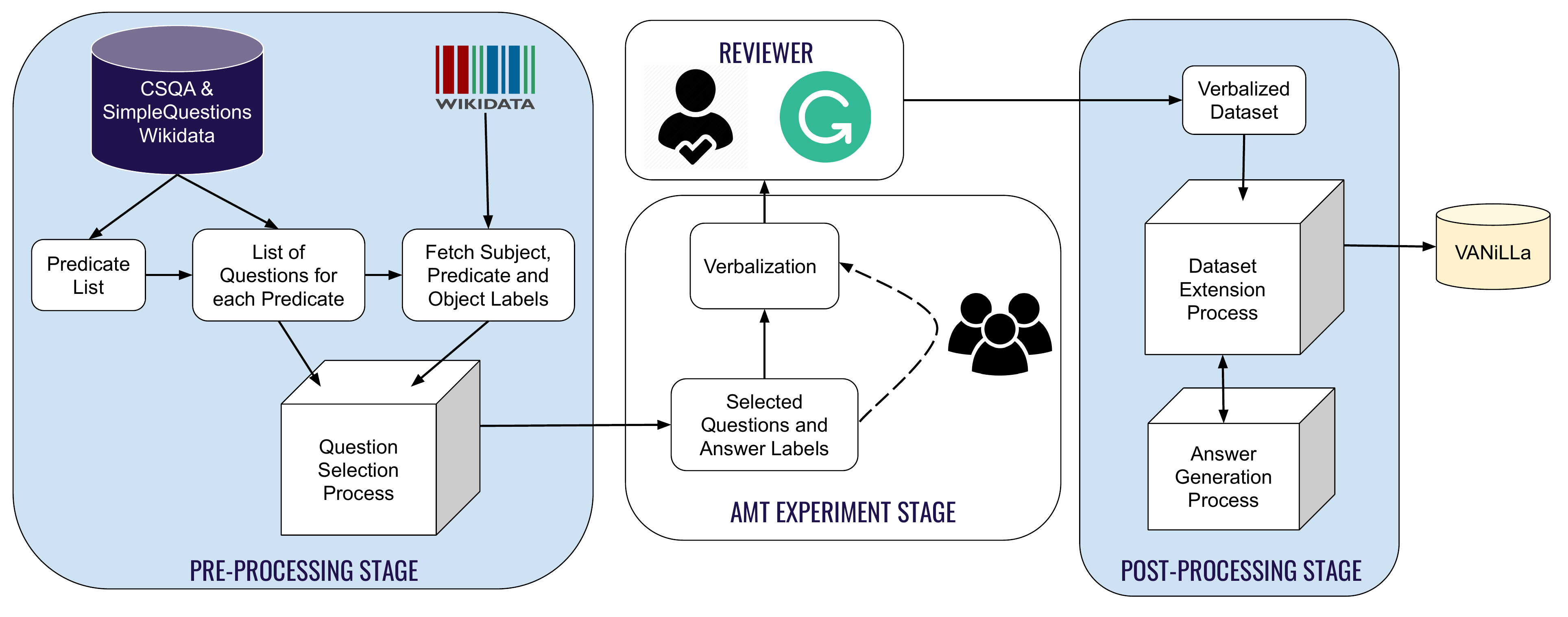}
\vspace{-15pt}
\caption{Dataset Generation Workflow} \label{fig2}
\end{figure}

Our dataset, \textbf{VANiLLa} (\textbf{Verbalized Answers in Natural Language at Large scale}) consists of simple questions $Q_{NL}$, selected from the CSQA~\cite{1801.10314} and SimpleQuestionsWikidata~\cite{wikidata-benchmark} datasets and answers $A_R$, labels fetched from the Wikidata~\cite{42240} Knowledge Graph, along with their answers in natural language $A_V$, generated using Amazon Mechanical Turk (AMT). We adapt our dataset from conventional Question Answering over Knowledge Graph (KGQA) datasets on Wikidata for the purpose of offering the possibility of extending them with verbalizations. The dataset creation process workflow uses a semi-automatic framework, as described in Figure
~\ref{fig2}, consists of four stages: \begin{inparaenum}[(i)]
 \item Pre-Processing Stage, 
 \item AMT Experiment Stage,
 \item Reviewer, and 
 \item Post-Processing Stage.
\end{inparaenum} 

\vspace{-10pt}

\subsubsection{Preprocessing Stage}
In this stage, we first collect all simple questions from CSQA\footnote{Note: We choose only direct questions where the KG entity directly exists in the question and no co-referencing is required.} and SimpleQuestionsWikidata datasets. While filtering questions, we ignore questions which produce more than one entity as an answer. This filtering process also generates a list of predicates present in the question for each dataset. On the basis of these predicates, we group the questions for a particular predicate to form a list. We then fetch labels for entities and relations present in the question and the answer from Wikidata. These labels along with the question lists are sent to the question selected process. The purpose of this pre-processing step is two fold:
\begin{inparaenum}[(i)]
    \item reducing the cost of crowd-sourcing the dataset from scratch, and 
    \item opening the possibility of extending the CSQA and SimpleQuestionsWikidata or SimpleQuestions dataset with the verbalized answers.
\end{inparaenum}

\subsubsection{Question Selection Process} 
The selection process starts by picking unique questions from the list of questions for a particular relation or predicate. 
Only a single instance of all the similar questions are selected in this process for the AMT experiment, thereby, reducing the cost of verbalization. Later, we may use the verbalizations of these selected questions as templates to automatically generate large scale data without the requirement of manual workforce. As illustrated in Algorithm~\ref{process}, the process takes $Q$ as input, which is a list of all filtered questions for a particular relation and outputs $S$, a list of selected questions for a particular relation. The first step is to copy the first question into the selected list $S$. Then we check for the 4-gram similarity between the other questions in list $Q$ with all the selected questions in list $S$ and continue adding dissimilar questions into list $S$, generating patterns of questions for each predicate.

\vspace{-13pt}
\begin{algorithm}
\caption{VANiLLa: Question Selection Process}
\label{process}
\hspace*{\algorithmicindent} \textbf{Input:} $Q = \{Q_1, Q_2,...., Q_n\}$ \\
\hspace*{\algorithmicindent} \textbf{Output:} $S = \{S_1, S_2,....,S_m\}$
\begin{algorithmic}[1]
\STATE $S \gets Q_1$
\FORALL{$Q_i \in Q$} 
\STATE {Check 4-gram similarity between $Q_i$ and $S_i$ $\forall S_i \in S$}
\IF{$Q_i \,and \,S_i\, are\, not\, similar\, \forall\, S_i \in S$}
\STATE {Add $Q_i$ to $S$}
\ENDIF
\ENDFOR
\end{algorithmic}
\end{algorithm}
\vspace{-13pt}

To check the similarity between two questions, we first mask the entities in both the questions and then find 4-grams for both questions. If any of the 4-grams match, then the questions are similar, otherwise they are dissimilar. For example, $S_1$ is \textit{"Who is the wife of Barack Obama?"} and $Q_1$ is \textit{"Who is the wife of Donald Trump?"}. The 4-grams for $S_1$ and $Q_1$ are \textit{\{"who", "is", "the", "wife"\}} and \textit{\{"is", "the", "wife", "of"\}}. Hence, $S_1$ and $Q_1$ are similar. Now, if we check $S_1$ with another question say $Q_2$, which is \textit{"Who is married to Barack Obama?"}. The 4-grams for $Q_2$ are \textit{\{"who", "is", "married", "to"\}} and do not match with $S_1$ . Hence, they are dissimilar. Therefore, $Q_2$ is added into the selected list $S$.

\subsubsection{AMT Experiment Stage}

In this stage, the human workers or turkers working on AMT are provided with a question $Q_{NL}$ from the selected list $S$ and its answer label $A_R$ retrieved from Wikidata~\cite{wikidata-benchmark}. The workers are instructed to rewrite the answers in natural language sentences so as to generate the answer sentence $A_V$ without re-writing the answer label. Thus, the crowd-sourcing task aims to verbalize the answers in natural language by incorporating the context of question without seeing the knowledge graph triples. The turkers are provided with explicit directions to verbalize the answer using the context of the question $Q_{NL}$. They are also presented with adequate examples to understand the task well.

\subsubsection{Reviewer}
A manual reviewing step is implemented to check for grammatical errors and invalid answers verbalized in the AMT stage. One native English speaking reviewer along with the help of the grammar checking tool\footnote{\href{https://app.grammarly.com/}{https://app.grammarly.com/}}, made the necessary corrections.

\subsubsection{Post-Processing Stage}
The last stage of the workflow yields a cost-effective extension of the dataset by utilizing the set of verbalizations received from the previous stage. The \textbf{Dataset Extension Process} searches for all similar question for every question in the verbalized dataset using 4-gram similarity and then calls the \textbf{Answer Generation Process} which in turn is employing the verbalized answer from the AMT experiment as a template to generate the answer in natural language for all the other similar questions. Due to these actions, we can provide a large-scale dataset with minimal cost.

\subsection{Dataset Characteristics}



Our dataset \textbf{VANiLLa} consists of a total of 107166 simple questions, with 85732 in training set and 21434 in test set. One significant feature of our dataset is that it is KG-independent and hence flexible for training verbalization models for any KG. VANiLLa covers a wide range of relations (300+) providing a significant amount of variations to the dataset. The different tags of the dataset are:
\begin{itemize}
    \item \textit{question-id}: an unique identification number for a dataset instance
    \item \textit{question}: question $Q_{NL}$ 
    \item \textit{answer}: retrieved answer $A_R$
    \item \textit{answer\_sentence}: verbalized answer in natural language $A_V$
\end{itemize}



However, some drawbacks of our dataset include the non-availability of multiple answer type questions, complex questions and paraphrased questions. 

\subsection{Reusability}
Our dataset can be used in multiple research areas. The straight forward utilization is in the domain of KGQA pipeline where the final representation of the answer can be modified into a natural language sentence. This also applies for Dialog systems where the user not only expects more human-like conversation but also a scope for answer verification from the question's point of view and for continuing the conversation. The dataset is KG independent, hence there is a possibility of using it for text based QA systems as well to generate answers in natural language. Also, our semi-automatic framework can be used for a cost-effective extension of the dataset. In addition, with some further investigation our dataset can be adopted in the form of multi-turns for assisting conversation type QA systems.

\subsection{Availability and Sustainability}
We have published the dataset at figshare\footnote{\href{https://figshare.com/articles/Vanilla\_dataset/12360743}{https://figshare.com/articles/Vanilla\_dataset/12360743}} under CC BY 4.0\footnote{\href{https://creativecommons.org/licenses/by/4.0/}{https://creativecommons.org/licenses/by/4.0/}} license to support sustainability. Figshare ensures data persistence and public availability, whereby guaranteeing all time accessibility of the dataset irrespective of the running status of our servers. The figshare project includes the training and the test sets of our dataset. 

Our baseline models and data creation framework is readily available as a open source repository\footnote{\href{https://github.com/AskNowQA/VANiLLa}{https://github.com/AskNowQA/VANiLLa}} under a GPL 3.0\footnote{\href{https://www.gnu.org/licenses/gpl-3.0.html}{https://www.gnu.org/licenses/gpl-3.0.html}} license. We intend to actively track feature requests and bug reports using our Github repository. 




\section{Potential Impact and Related Work} \label{Sec:Litrature}
To understand the impact of our dataset we first 
showcase the
previous works that have been done in this domain. 
We report some of the resources which are currently available for verbalizing answers. GENQA \cite{yin-etal-2016-neural-generative} and COREQA \cite{he-etal-2017-generating} are two datasets in Chinese language which provide answers in the form of sentences rather than a concise answer. To the best of our knowledge, VQuAnDa \cite{kacupaj2020vquanda} is the first and only dataset in English language to provide answer verbalization. The dataset contains questions from LC-QuAD \cite{DBLP:conf/semweb/TrivediMDL17} on DBpedia as the target KG. The advantage of this dataset is doubled due to the availability of SPARQL representation for questions. Thus, it is also useful for model's verbalizing answers from a SPARQL query. The dataset was aimed to enable users to validate the information generated from QA systems by incorporating not only the retrieved information but also additional characteristics indicating the process of answer retrieval.  Nevertheless, this dataset does not cover key features of the user question to mimic human conversation in addition to enabling answer validation. 

VANiLLa is a large scale dataset with over 100k question answer pairs compared to the 5k pairs in the VQuAnDa dataset. VQuAnDa  aims to generate verbalizations using the SPARQL query and find the position of the answer in the natural language sentence hence in return has to implement a slot filling approach to achieve the final response. 
Whereas VANiLLa promotes an end-to-end approach by avoiding any post processing of the generated response. 
The VQuAnDa dataset generation requires a SPARQL query, whereas the VANiLLa generation workflow only requires answers. This showcases that Vanilla’s architecture is KG-independent, and one could extend the VANiLLa dataset generation workflow to a QA dataset with no KG or formal query in the background. 
In this regard, our dataset would impact the following sectors of the NLP community:
\begin{itemize}
        \item[\textbf{Natural Language Generation:}]     Popular NLG tasks and dataset on RDF verbalzation, such as WebNLG\cite{colin-etal-2016-webnlg}, do not take into consideration the question elements when reconstructing the facts in natural language. Hence, the sentences generated involve relevant information from the facts without any context of the question. From this standpoint, our dataset would open up new paradigm in NLG research.
    \vspace{10pt}
    \item[\textbf{Question Answering and Dialog Systems:}] 
        Different QA datasets based on Text, such as SQuAD \cite{rajpurkar2016squad} and NaturalQuestions \cite{47761}, and KG, such as SimpleQuestion \cite{bordes2015largescale}, WebQuestionSP \cite{yih-etal-2016-value}, ComplexWebQuestions \cite{talmor2018web}, QALD-9 \cite{inproceedings3}, LC-QuAD 1.0 \cite{DBLP:conf/semweb/TrivediMDL17}, LC-QuAD 2.0 \cite{inbook}, exists. Furthermore, datasets are also adapted into conversation or dialog format with multiple utterances such as, QuAC \cite{choi2018quac}, CoQA \cite{reddy2019coqa}, CSQA \cite{1801.10314}, SQA \cite{iyyer-etal-2017-search}. Recent trend in QA datasets are based on the following aspects:
        \begin{inparaenum}[(1)]
            \item expanding the size of dataset,
            \item extending the dataset with more variations in question type such as aggregations, boolean queries, etc.,
            \item enhancing question complexity using compound features like comparison or unions,
            \item enhancements covering multiple utterances in the form of conversations.
        \end{inparaenum} 
\end{itemize}

At present, the datasets associated with the task of QA or dialog systems provide concise or to-the-point answers for questions posed by users. However, verbalization of answers, which enable a more human-like response by incorporating additional information, are neglected in these datasets. 
From this viewpoint, VANiLLa will provide a more contextual and human-like response to user questions, using elements from the question to generate the verbalized answers. Furthermore, it is beneficial in dialog systems to have a continuous flow in the conversation, by enabling users with more content for the next utterance.


\section{Evaluation} \label{Sec:eval}
\vspace{-5pt}
In this section, we report the different experiments we have performed for the assessment of the quality of our dataset and for comparing the performance of our baseline models.

\subsection{Experiment: 1} \label{Sec:Exp1}

\subsubsection{Objective:} A more human-like response to a user question $Q_{NL}$ would mean that the generated response in natural language provides some context of the question. From this point of view, we perform this Experiment  to show that the verbalized answers $A_V$ in our dataset preserves the semantic and syntactic meaning of the question more compared to the relevant RDF triples verbalization $A_T$. Additionally, we also indicate that a response generated by verbalizing the RDF triple $A_T$ would be semantically and syntactically more close to the $T$ in contrast to the that of $A_V$ and $T$. Concisely, we evaluate hypothesis \textbf{H1} from section: \ref{hyp}.

\subsubsection{Experimental Settings:} We randomly sample 500 questions from our dataset to perform this experiment. For similarity comparison purposes with the natural language answers in our dataset and user question, we generated the verbalization of the RDF triples associated with these questions with the help of a simple AMT experiment where the turkers were asked to rewrite the triples in the form of natural language sentences without any information about the questions. 

To analyze the quality of our dataset, we calculate the similarity measure $Sim$: \begin{inparaenum} [(i)]
    \item  as the Levenshtein distance using Fuzzywuzzy, denoting the "syntactic similarity" and
    \item as the cosine vector similarity using BERT embeddings, denoting the "semantic similarity",  
\end{inparaenum} of $T$ and $Q_{NL}$ with both $A_V$ and $A_T$. We record the mean $\mu$ and standard deviation $\sigma$ for all the comparisons over the randomly sampled data.


\begin{figure} [h]
    \centering
    \begin{tabular}{c}
    \includegraphics[width = \textwidth]{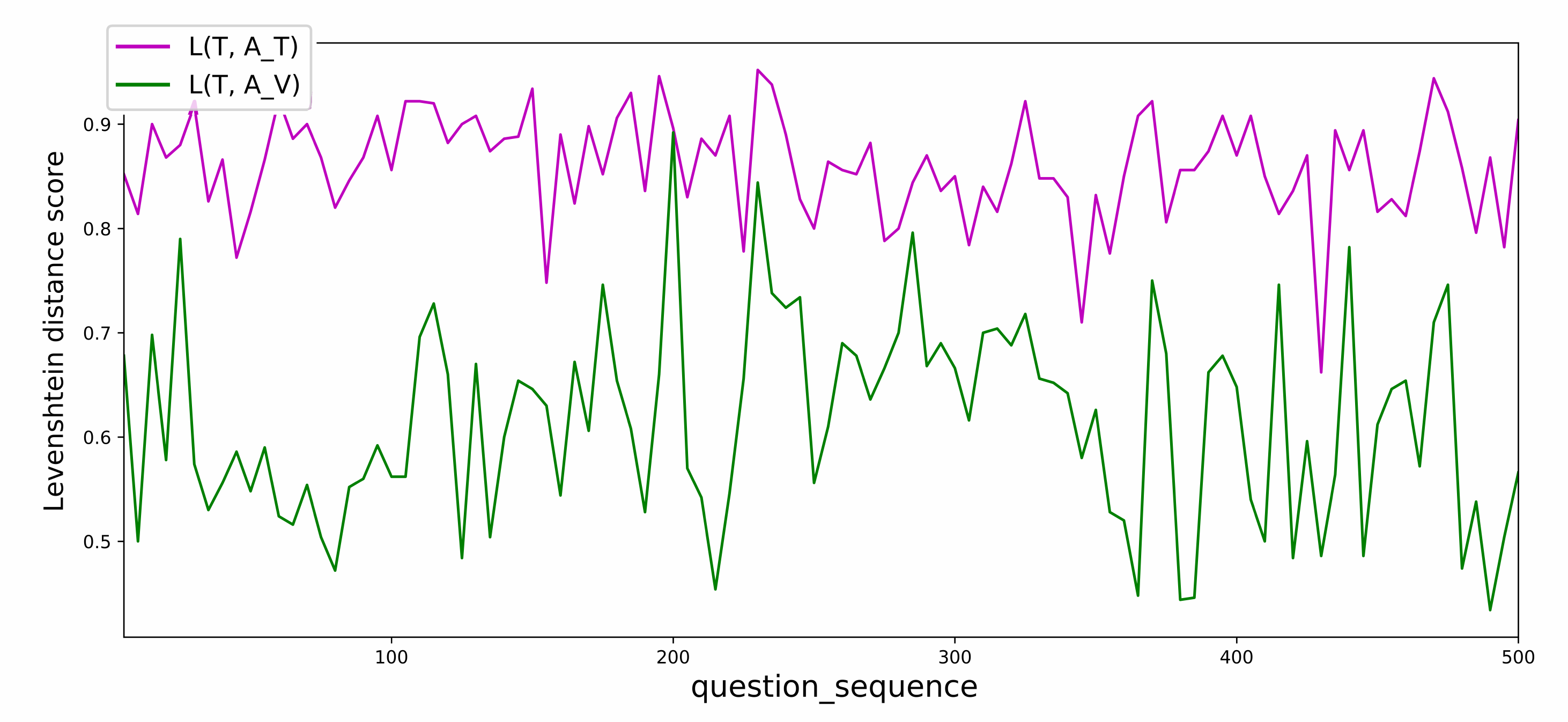}   \\
    \includegraphics[width = \textwidth]{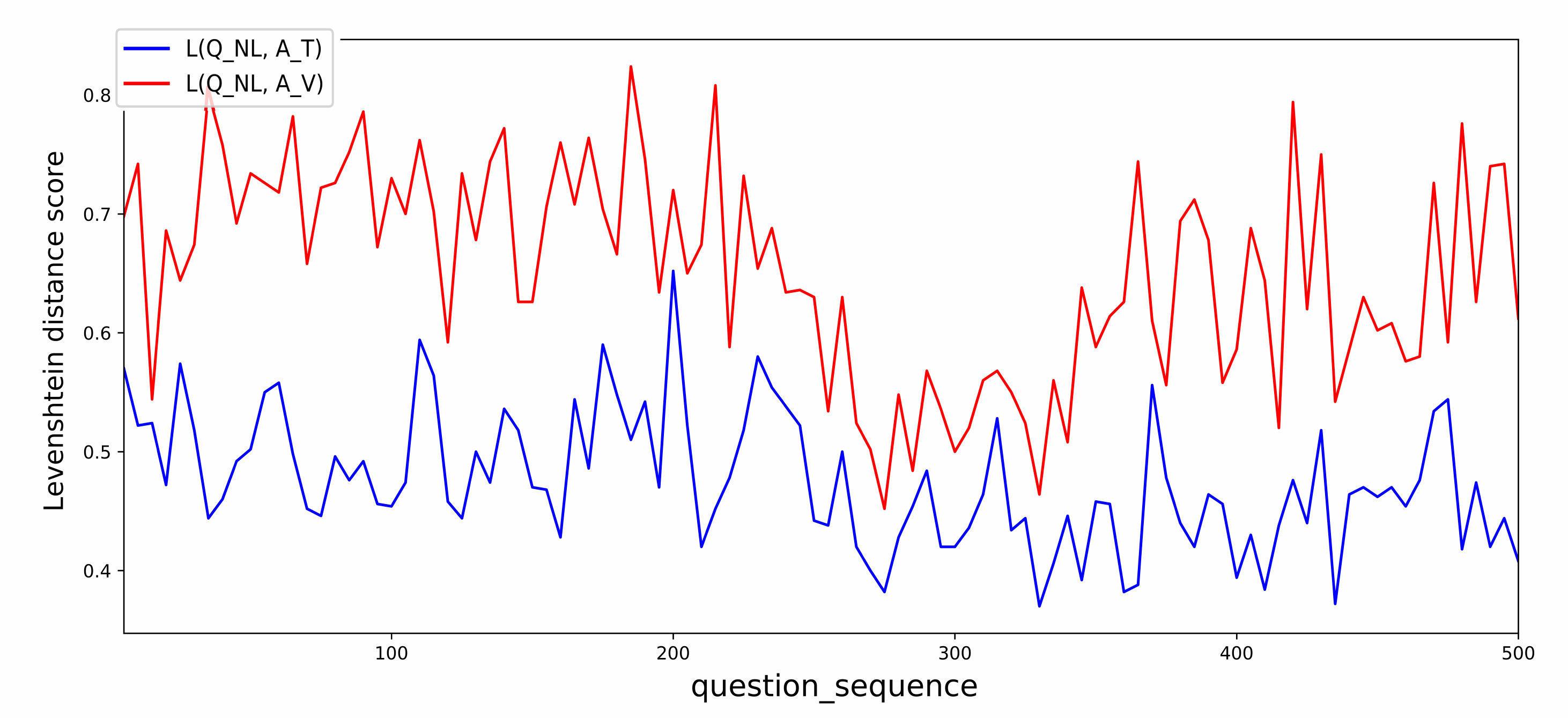}    
    \end{tabular}
    \vspace{-5pt}
    \caption{Syntactic and Semantic Comparison of $T$ and $Q_{NL}$ with both $A_V$ and $A_T$ }
    \label{exp1}
\end{figure}

\subsubsection{Experimental Results:}
The results of our experiment are depicted in Table: \ref{tab:exp1}. On the basis of Levenshtein distance, we observe that the syntactic similarity between the verbalized RDF triple $A_T$ and the RDF triple $T$ is more than the similarity between $T$ and the verbalized answer $A_V$, i.e., $Sim(A_T, T) > Sim(A_V, T)$. In contrary, the syntactic similarity between $A_V$ and the user question $Q_{NL}$ is more than the similarity $Q_{NL}$ and the verbalized RDF triple $A_T$, i.e., $Sim(A_V, Q_{NL}) > Sim(A_T, Q_{NL})$. We also observe the same in the case of the cosine similarity measure.
Thus, empirically proving hypothesis \textbf{H1}.

\begin{table} [h]
\centering 
     \begin{tabular}{@{}ccccccccc@{}} 
    \toprule 
    & & & & \multicolumn{2}{c}{\textbf{Levenshtein Ratio}}  & & \multicolumn{2}{c}{\textbf{Cosine-Similarity} }                 \\
    \cmidrule{5-6}\cmidrule{8-9}
    & & & & $\mathbf{A_{T}}$  & $\mathbf{A_{V}}$ & & $\mathbf{A_{T}}$  & $\mathbf{A_{V}}$
      \\
    \midrule
    \multirow{2}{*}{$\mathbf{T}$} & & $\mu$ & & 0.8602 & 0.6139 
    & &  0.8441  & 0.7965  \\
     & & $\sigma$ & & 0.1066 &  0.1779
    & &  0.0793 & 0.0992 \\
    \midrule
    \multirow{2}{*}{$\mathbf{Q_{NL}}$} & & $\mu$ & & 0.4750  & 0.6519 
    & &  0.3528  & 0.3574  \\
     & & $\sigma$ & & 0.1018 & 0.1504
    & &  0.1064 & 0.1056 \\
    \bottomrule
    \end{tabular}
    \vspace{10pt}
    \caption {Comparing $A_V$, $A_T$, $Q_{NL}$ and $T$ based on semantic cosine vector and syntactic Levenshtein distance similarity}
    \label{tab:exp1}
\end{table}





Additionally, Figure: \ref{exp1} illustrates the results of the experiment calculating the Levenshtein distances. In the first plot, we compare the Levenshtein distance between ($T$, $A_T$) and ($T$, $A_V$) over the 500 randomly sampled questions. The plot clearly depicts that the purple line indicating the distance between ($T$, $A_T$) is higher than the green line indicating the distance between ($T$, $A_V$) at most data point, thus demonstrating that the pair ($T$, $A_V$) is syntactically less similar than the ($T$, $A_T$). Furthermore in the second plot, we compare the Levenshtein distance between ($Q_{NL}$, $A_T$) and ($Q_{NL}$, $A_V$), which clearly depicts the red line indicating the distance between ($Q_{NL}$, $A_V$) is higher than the blue line indicating the distance between ($Q_{NL}$, $A_T$). Thus, verifying that $A_V$ is syntactically more similar to $Q_{NL}$ than $A_T$. However, in both case there are a few points of exception where the similarity measures are the same. This is due to questions in the dataset which contain same surface labels as in the RDF triples.

\subsection{Experiment: 2}

\vspace{-2pt}
\subsubsection{Objective:}
The objective of this experiment is to record the performance of our baseline models on VANiLLa dataset based on certain evaluation criteria.

\vspace{-5pt}
\subsubsection{Baseline models:}
This section presents an overview of our baseline models used for evaluating and assessing the quality of the data. We decided to use some conventional sequence-to-sequence models following the underlying Encoder-Decoder pipeline \cite{sutskever2014sequence,cho2014learning}.
The question $Q_{NL}$ and the retrieved answer $A_T$ are separated with the separation "$<$\textit{sep}$>$" tag to form the input to the encoder and the decoder produces the verbalized answer in natural language $A_V$ as the output. The inputs to both the encoder and the decoder end with the end-of-sequence "$<$\textit{eos}$>$" tag and the output from the decoder is appended with the start-of-sequence "$<$\textit{sos}$>$" tag before the start of prediction. The four baseline models are as follows:
\begin{itemize}
    \item[\textbf{Seq2Seq model with attention mechanism:}]
        Sequence-to-Sequence model with attention mechanism \cite{bahdanau2014neural} proposed by Bahdanau simultaneously learns to align and translate by employing a bi-directional RNN as the encoder and using a weighted sum of the vector representation generated at the encoder level as the context in the decoder. These sequence-to-sequence models have been widely used in the NLP community for tasks such as Machine Translation, Question Answering, etc.
    \vspace{10pt}
    \item[\textbf{Convolution based Encoder-Decoder model:}]
        Facebook AI introduced the use of CNN in sequence learning \cite{gehring2017convolutional}. The concept is to implement a fully convolutional architecture, by incorporating CNN at both encoder and decoder level. These models implement a position embedding, which provides an understanding about the portion of input or output sequence currently in consideration, along with a multi-step attention. 
    \vspace{10pt}
    \item[\textbf{Transformer:}]
        Transformers are novel architectures that solves sequence-to-sequence tasks quite nicely because of its ability to handle long-term dependencies. Recently, Transformers achieved state-of-the art in wide range of NLP tasks such as Natural Language Generation, Machine Translation. The architecture of transformer was first proposed by Vaswani et al.~\cite{vaswani2017attention}. Followed by that many pre-trained transformer models are proposed such as BERT~\cite{devlin2018bert}, GPT~\cite{radford2018improving}, RoBERTa~\cite{liu2019roberta}, which are trained on a large number of data. For simplicity, we use a simple transformer model for the evaluation of our dataset. Multi-headed attention and positional-encoding are used in this transformer model. Multi-headed attention helps to capture global dependencies between input and output and positional-encoding encodes information about the relative or absolute position of each token in the sequence.
\end{itemize}

\vspace{-5pt}
\subsubsection{Dataset:}
We use our dataset VANiLLa with a 80-20 split to generate the training and test sets. 
The training set consists of 85732 examples whereas the test set consists of 21434 examples. We also implement a $k$-cross validation with $k=5$ on training set for generating the validation set in both cases.

\vspace{-5pt}
\subsubsection{Experimental Settings:}
We use standard dropout value of 0.5 for the recurrent layers, 0.25 for convolution layers and 0.1 for multi-head attention mechanism. The hidden layers dimensions for transformers is set to 200 with embedding dimension of 200, while for the other baselines are set the hidden dimension to 512 and embedding dimension to 300. In the case of Convolution based Encoder Decoder model, we utilize 10 layers of convolutions with kernel size 3. For the multi-headed attention mechanism in Transformer model, we set 8 as the number of head with 4 layers. For the sake of consistency, we train all the models for 
7 epochs on the our dataset, with Adam optimizer and cross entropy loss function. 

\vspace{-5pt}
\subsubsection{Evaluation Criteria:}
We implement the following criteria for the evaluation and comparison of our baseline models: 
\begin{itemize}
    \item[Criteria\#1] \textbf{Perplexity:} Perplexity defines the quality of predicting real sentences for a probability model. A good model will have a lower perplexity indicating a high probability of prediction. It is the average branching factor for predicting the next word. It is calculated with the help of cross-entropy as follows:
    
    \begin{equation}
        PPL = 2^{H(p,q)}
    \end{equation}
    where, $H(p,q)$ is the cross-entropy of the model which tries to learn the probability distribution $q$ close to the unknown probability distribution of the data $p$.
    \newline
    \item[Criteria\#2] \textbf{Modified Precision:} Modified Precision is the sum of the clipped n-gram counts of the generated sequence or hypothesis divided by the total number of n-grams in the hypothesis. The range of the modified precision is between 100 and 0 with 100 being the best score. This scores depicts two aspects: 
    \begin{inparaenum} [(i)] \item adequacy (generated hypothesis sentence contains similar words as that in the reference sentence), and 
    \item fluency (the n-gram matches between the hypothesis and reference).
    \end{inparaenum}
    \newline
    \item[Criteria\#3] \textbf{BLEU:} The BLEU score, Bilingual Evaluation Understudy, is a modified precision score \cite{10.3115/1073083.1073135} combined with a brevity penalty for penalizing shorter generation. Hence, the BLEU score checks for matches in length, order and word choice between the hypothesis and the references. The range of BLEU score is between 100 and 0 with 100 being the best score. 
\end{itemize}

\begin{figure}[h]
\centering
\includegraphics[width=0.9\textwidth]{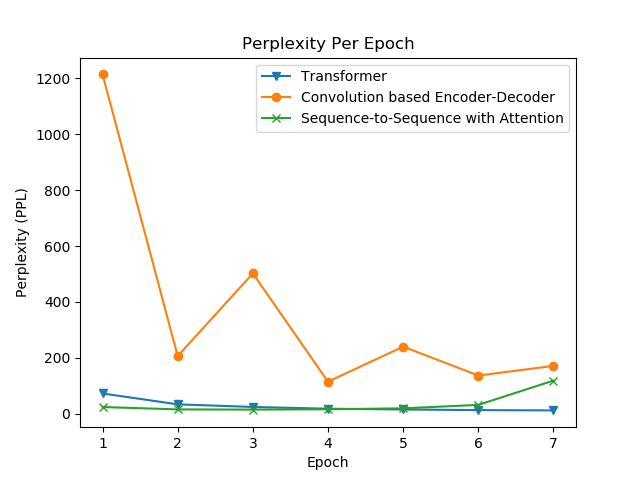}

\caption{Perplexity(PPL) per Epoch during training} \label{ppl}
\end{figure}
\vspace{-20pt}

\begin{table} [h]
\centering 
     \begin{tabular}{@{}ccccc@{}} 
    \toprule 
      \textbf{Baseline Model}  & & \textbf{PPL}  &\textbf{Precision}  &\textbf{BLEU}  
      \\
    \midrule
    Seq2Seq with attention & &  27.91 & 19.84 & 16.66 \\
    CNN Enc-Dec & & 87.67 & 70.50 & 15.42\\
    Transformer & & \textbf{12.10} & \textbf{76.00} &  \textbf{30.80} \\
    \bottomrule
    \end{tabular}
    \vspace{3mm}
    \caption {Evaluating Baseline Models on VANiLLa Dataset (No. of Epochs: 7)}
    \label{exp2}
\end{table}
\vspace{-20pt}

\subsubsection{Experimental Results:}
The perplexity value on the validation set after every epoch for all our baselines are plotted in Figure: \ref{ppl} and the test results of the three baseline models trained on the two sizes of our dataset, VANiLLa has been reported in table: \ref{exp2}. The results clearly indicate that the transformer model performs the best among the three baselines with perplexity value of 12.10 and BLEU score of 30.80.

The transformer model puts attention on the whole sentence instead of attention on word by word, hence capturing the context better. Additionally, transformer handles the relationship between words in a sentence by positional encoding and multi-headed attention mechanism \cite{vaswani2017attention}. These characteristics helps transformer model achieve better result than other baseline models.

\subsection{Experiment: 3}

\subsubsection{Objective:} Empirically it is evident that the transformer model is the best amongst the baseline models. Hence, in this experiment we examine the improvement of the evaluation metrics for the transformer model trained on the combination of VANiLLa and VQuAnDa~\cite{kacupaj2020vquanda} datasets (5000 question-answer pairs).
\begin{table} [h]
\centering 
     \begin{tabular}{@{}ccccccc@{}} 
    \toprule 
      \textbf{Training Set}  & & \textbf{Test Set} & & \textbf{PPL} &\textbf{Precision}  &\textbf{BLEU}  
      \\
    \midrule
    \multirow{2}{*}{VANiLLa + VQuAnDa} & &  VQuAnDa & & 132.26 &  \textbf{80.45} & 21.61  \\
     & & VANiLLa & &  9.05 & 73.63 & \textbf{35.70} \\
    VQuAnDa & & VQuAnDa & & 84.88 & 75.05 & 26.25 \\
    \bottomrule
    \end{tabular}
    \vspace{3mm}
    \caption {Evaluating Transformer Model on VANiLLa and VQuAnDa Dataset (No. of Epochs 50)}
    \label{exp3}
\end{table}
\vspace{-30pt}

\subsubsection{Experimental Settings:} In this experiment we combine the training data of VANiLLa with the training data of VQuAnDa dataset and feed it into the transformer model. Although, we combine the training data for the training phase but we test the model on both the VANiLLa and VQuAnDa separately. Note that the majority of the combined data is from VANiLLa as the number of data in VANiLLa is 20 times more than VQuAnDa. 

\subsubsection{Experimental Results:}
 The combined data used for training improves the precision score (80.45) of VQuAnDa test data by a fair margin. Additionally, transformer model achieves best BLEU score on VANiLLa test set. So, it is evident that this combined data training approach helps in creating a strong baseline for future research. Furthermore, our baseline transformer model with the suggested parameters (described in the experimental settings) achieve SOTA in BLEU score on VQuAnDa data which is 26.25 where in VQuAnDa paper BLEU is reported 18.18.


\section{Conclusion and Future Work}
 We introduce the VANiLLa dataset with the intent to support the verbalization of answers in the KGQA task. The dataset consisting of over 100k examples was generated semi-automatically by adapting questions from other datasets such as CSQA and SimpleQuestionsWikidata and then verbalizing the answers by crowd-sourcing. We empirically show that the verbalizations in our dataset are syntactically and semantically closer to the question than to the KG triple labels. Additionally, we evaluate multiple baseline models based on generic sequence to sequence models already in use in the NLP community. These baseline models present a reference point for future research. The experimental results of these baselines indicate that there still exists substantial room for enhancement for solving the task of verbalizing answers. 
 
 One major use case of answer verbalizations are dialogue systems. We believe that our dataset along with our design framework presents a valuable contribution in this field. In the future, we plan to continuously aid researchers with more variations and expansions of our dataset. 
\bigskip 

\bibliographystyle{abbrv}
\bibliography{reference}

\appendix 

\end{document}